\documentclass{article}
\usepackage{amsmath, amsfonts}
\usepackage{times}
\usepackage{graphicx} 
\usepackage{subfigure} 

\usepackage{natbib}

\usepackage{algorithm}
\usepackage{algorithmic}

\usepackage{hyperref}

\usepackage[accepted]{icml2017} 

\icmltitlerunning{Biologically inspired protection of deep networks from adversarial attacks}

\newcommand{\Wl}[1]{{\mathbf{W}}^{#1}}
\newcommand{\w}{\mathbf{w}}
\newcommand{\x}{\mathbf{x}}

\newcommand{\h}{\mathbf{h}}
\newcommand{\bi}{\mathbf{b}}
\newcommand{\y}{\mathbf{y}}

\begin{document} 

\twocolumn[
\icmltitle{Biologically inspired protection of deep networks from adversarial attacks}



\icmlsetsymbol{equal}{*}

\begin{icmlauthorlist}
\icmlauthor{Aran Nayebi}{stn}
\icmlauthor{Surya Ganguli}{st}
\end{icmlauthorlist}

\icmlaffiliation{stn}{Neurosciences PhD Program, Stanford University}
\icmlaffiliation{st}{Department of Applied Physics, Stanford University}

\icmlcorrespondingauthor{Aran Nayebi}{anayebi@stanford.edu}
\icmlcorrespondingauthor{Surya Ganguli}{sganguli@stanford.edu}

\icmlkeywords{boring formatting information, machine learning, ICML}

\vskip 0.3in
]



\printAffiliationsAndNotice{} 

\begin{abstract} 

Inspired by biophysical principles underlying nonlinear dendritic computation in neural circuits, we develop a scheme to train deep neural networks to make them robust to adversarial attacks. Our scheme generates highly nonlinear, saturated neural networks that achieve state of the art performance on gradient based adversarial examples on MNIST, despite never being exposed to adversarially chosen examples during training. Moreover, these networks exhibit unprecedented robustness to targeted, iterative schemes for generating adversarial examples, including second-order methods. We further identify principles governing how these networks achieve their robustness, drawing on methods from information geometry. We find these networks progressively create highly flat and compressed internal representations that are sensitive to very few input dimensions, while still solving the task. Moreover, they employ highly kurtotic weight distributions, also found in the brain, and we demonstrate how such kurtosis can protect even linear classifiers from adversarial attack. 
\end{abstract} 

\section{Introduction}
Deep Neural Networks (DNNs) have demonstrated success in many machine learning tasks, including image recognition \cite{imagenet}, speech recognition \cite{speech}, and even modelling mathematical learning \cite{deepknowledge}, among many other domains. However, recent work has exposed a remarkable weakness in deep neural networks \cite{szegedy} (see \cite{survey} for a survey), namely that very small perturbations to the input of a neural network can drastically change its output. In fact, in image classification tasks, it is possible to perturb the pixels in such a way that the perturbed image is indistinguishable from its original counterpart to a human observer, but the network's class prediction is completely altered. These adversarial examples suggest that despite the above successes, machine learning models are not fundamentally understanding the tasks that they are trained to perform. 

Furthermore, the imperceptibility of these adversarial perturbations to human observers suggests that these machine learning algorithms are performing computations that are vastly different from those performed by the human visual system.  This discrepancy is of particular scientific concern as deep neural networks now form foundational models in neuroscience for the visual processing stream  \cite{yamins, deepretina, kreview}. So their susceptibility to adversarial perturbations that are imperceptible to us suggest our models are missing a fundamental ingredient that is implemented in the brain. However, the existence of adversarial examples is also of particular technological concern in machine learning, as these adversarial examples generalize across architectures and training data, and can therefore be used to attack machine learning systems deployed in society, without requiring knowledge of their internal structure \cite{papernot0, survey}. 

It is important to note that adversarial examples of this form are not limited to deep networks but are also an issue even in linear high dimensional classification and regression problems. A plausible explanation \cite{goodfellow} for the existence of these adversarial examples lies in the idea that any algorithm that linearly sums its high dimensional input vectors with many small weights can be susceptible to an attacker that adversarially perturbs each of the individual inputs by a small amount so as to move the entire sum in a direction that would make an incorrect classification likely. This idea lead to a fast method to find adversarial examples which could then be used to explicitly train neural networks to be robust to their own adversarial examples \cite{goodfellow}.
 
However, it is unclear that biological circuits explicitly find their own adversarial examples by optimizing over inputs and training against them.  Therefore, we are interested in guarding against adversarial examples in a more biologically plausible manner, without explicitly training on adversarial examples themselves. Of particular interest is isolating and exploiting fundamental regimes of operation in the brain that prevent the imperceptible perturbations that fool deep networks, from fooling us.  In this paper, we take inspiration from one fundamental aspect of single neuron biophysics that is not often included in artificial deep neural networks, namely the existence of nonlinear computations in intricate, branched dendritic structures \cite{dendrites,biophysicscomp,londonrev}.  These nonlinear computations prevent biological neurons from performing weighted sums over many inputs, the key factor thought to lead to susceptibility to adversarial examples.  Indeed, the biophysical mechanism for linear summation in neurons corresponds to the linear superposition of trans-membrane voltage signals as they passively propagate along dendrites. These voltage waves can linearly sum synaptic inputs. However, there is also a high density of active ionic conductances spread through the dendritic tree that can destroy this linear superposition property in purely passive dendrites, thereby limiting the number of synapses that can linearly sum to $O(10)-O(100)$.  These active conductances lead to high threshold, nonlinear switch like behavior for voltage signalling. As a result, many parts of the dendritic tree exist in voltage states that are either far below threshold, or far above, and therefore saturated. Thus biological circuits, due to the prevalence of active dendritic processing, may operate in a highly nonlinear switch-like regime in which it is very difficult for small input perturbations to propagate through the system to create large errors in output.

Rather than directly mimic this dendritic biophysics in artificial neural networks, here we take a more practical approach and take inspiration from this biophysics to train artificial networks into a highly nonlinear operating regime with many saturated neurons.  We develop a simple training scheme to find this nonlinear regime, and we find, remarkably, that these networks achieve state of the art robustness to adversarial examples despite never having access to adversarial examples during training.  Indeed we find 2-7\% error rates on gradient-based adversarial examples generated on MNIST, with little to no degradation in the original test set performance. 

Furthermore, we go beyond performance to scientifically understand which aspects of learned circuit computation confer such adversarial robustness.  We find that our saturated networks, compared to unsaturated networks, have highly kurtotic weight distributions, a property that is shared by synaptic strengths in the brain \cite{lognormal}. Also, our networks progressively create across layers highly clustered internal representations of different image classes, with widely separated clusters for different classes.  Furthermore we analyze the information geometry of our networks, finding that our saturated networks create highly flat input-output functions in which one can move large distances in pixel space without moving far in output probability space.  Moreover, our saturated networks create highly compressed mappings that are typically sensitive to only one direction in input space.  Both these properties make it difficult even for powerful adversaries capable of iterative computations to fool our networks, as we demonstrate.  Finally, we show that the highly kurtotic weight distributions that are found both in our model and in biological circuits, can by themselves confer robustness to adversarial examples in purely linear classifiers. 

\section{Adversarial Example Generation}
We consider a feedforward network $F$ with $D$ layers of weights $\Wl{1},\ldots, \Wl{D}$ and $D+1$ layers of neural activity vectors $\x^0,\dots,\x^D$, with $N_l$ neurons in each layer $l$, so that $\x^{l}\in \mathbb{R}^{N_l}$ and $\Wl{l}$ is an $N_{l} \times N_{l-1}$ weight matrix.  The feedforward dynamics elicited by an input $\x^0$ are
\begin{equation*}
\begin{split}
\x^{l} &= \phi(\h^l) \quad \h^l = \Wl{l} \, \x^{l-1} + \bi^l \,\,\, \text{for} \,\,\,  l=1,\dots,D-1\\
\x^{D} &= \text{softmax}(\h^{D}),
\end{split}
\label{eq:netdynam}
\end{equation*}
where $\bi^l$ is a vector of biases, $\h^l$ is the pattern of inputs to neurons at layer $l$, and $\phi$ is a single neuron scalar nonlinearity that acts component-wise to transform inputs $\h^l$ to activities $\x^l$. We take $\y$ to be the class indicator vector generated from $\x^{D}$. We also denote by $\x^D = F(\x^0)$ the network's composite transformation from input to output. 

For such networks, the essential idea underlying adversarial examples is to start with a test example $\x^0$ that is correctly classified by the network with class indicator vector $\y$, and transform it through an additive perturbation $\Delta \x^0$ into a new input $\x^0 + \Delta \x^0$ that is incorrectly classified by the network $F$ as having a ``goal'' class label $\y^G \neq \y$. Moreover, the perturbation $\Delta \x^0$ should be of bounded norm so as to be largely imperceptible to a human observer. This idea leads naturally to an optimization problem:
\begin{equation}\label{eq1}
\arg\min_{\Delta \x^0}\|\Delta \x^0\|
\quad \text{s.t. }F(\x^0 + \Delta \x^0) = {\y}^G.
\end{equation}

However, as this is a complex optimization, many simpler methods have been proposed to efficiently generate adversarial examples (e.g. \cite{goodfellow,miyato, papernot0}). In particular, the fast gradient sign method of \citet{goodfellow} is perhaps the most efficient method. Motivated by the notion that adversarial attacks can arise even in linear problems in high dimensional spaces, \citet{goodfellow} linearized the input-output map $F$ around the test example and searched for bounded $l_\infty$ norm perturbations that maximize the network's cost function over the linearized network. More precisely, suppose the cost function of the network is $C_0 = C(F(\x^0), \y)$, then its linearization is 
\begin{equation}
C(F(\x^0+\Delta \x^0), \y) \approx C_0 + (\nabla_{F}C)J (\Delta \x^0),
\end{equation}
where $J$ is the Jacobian of $F$. Then the bounded $l_\infty$ norm optimization that maximizes cost has the exact solution
\begin{equation}\label{eq2}
\epsilon\text{sgn}(\nabla_{F}C J)
 = \arg\max_{\Delta \x^0} (\nabla_{F}C)J(\Delta \x^0)
 \text{ s.t. }\|\Delta \x^0\|_{\infty} \leq \epsilon.
\end{equation}
If a network can be susceptible to these gradient-based adversaries, then we can choose $\epsilon$ to be small enough for the given dataset so it is imperceptible to human observers yet large enough for the network to misclassify. For MNIST, \citet{goodfellow} took $\epsilon = 0.25$, since each pixel is in $[0, 1]$. We follow this prescription in our experiments.

With efficient methods of generating adversarial examples \eqref{eq2}, \citet{goodfellow} harnessed them to develop adversarial training, whereby the network is trained with the interpolated cost function:
\begin{equation}\label{eqadv}
\alpha C + (1-\alpha)C(F(\x^0 + \epsilon\text{ sgn}(\nabla_{F}C)J), \y),
\end{equation}
As a result, the network is trained at every iteration on adversarial examples generated from the current version of the model. On maxout networks trained on MNIST, \citet{goodfellow} found that they achieved an error rate of 89.4\% on adversarial examples, and with adversarial training (where $\alpha = 0.5$), they were able to lower this to an error rate of 17.9\%.

We now turn to ways to avoid training on adversarial examples, in order to have the networks be more intrinsically robust to their adversarial examples. \citet{papernot} suggested knowledge distillation, which involves changing a temperature parameter $T$ on the final softmax output in order to ensure that the logits are more spread apart. However, the authors do not try their approach on adversarial examples generated by the fast gradient sign method, nor does this approach address the broader criticism of \citet{goodfellow} that models susceptible to gradient-based adversaries operate heavily in the linear regime.  We develop a method that strongly departs from the high dimensional linear regime in which adversarial examples abound. The basic idea is to force networks to operate in a nonlinear saturating regime. 

\section{Saturating Networks}
A natural starting point to achieve adversarial robustness is to ensure that each element of the Jacobian of the model, $J = \partial F/\partial \x^0$, is sufficiently small, so that the model is not sensitive to perturbations in its inputs. Jacobian regularization is therefore the most direct method of attaining this goal; however, for sufficiently large networks, it is computationally expensive to regularize the Jacobian as its dimensions can become cumbersome to store in memory. 

An immediate alternative would be to use a contractive penalty as in \citet{gu}, whereby the Frobenius norm of the layer-wise Jacobian is penalized:
\begin{equation}
\sum_{l = 1}^{D}\lambda_l\left\|\frac{\partial \h^l}{\partial \h^{l-1}}\right\|_F,
\end{equation}
where each $\lambda_{l} \in \mathbb{R}$. For element-wise nonlinearities, \citet{cae} show that this penalty can be computed in $O\left(\max_{l}\left(\left|\x^l\right|\times \left|\x^{l-1}\right|\right)\right)$ time, where $\left|\cdot\right|$ denotes the length (number of units). 

While indirectly encouraging the activations to be pushed in the saturating regime of the nonlinearity, this contractive penalty can nonetheless be practically difficult to compute efficiently for networks with a large number of hidden units per layer, and also tends to limit the model's capacity to learn from data, degrading test set accuracy.

Saturating autoencoders were introduced by \citet{saturate} as a means of explicitly encouraging activations to be in the saturating regime of the nonlinearity, in order to limit the autoencoder's ability to reconstruct points that are not close by on the data manifold. Their penalty takes the following form for a given activation $\h = \mathbf{W}\x + \bi$ and $\lambda\in\mathbb{R}$,
\begin{equation}
\begin{split}
\lambda\sum_{i = 1}^{|\h|}\phi_c({\h}_i),
\end{split}
\end{equation}
where the complementary function is defined as:
\begin{equation}\label{comp}
\phi_c(z) \equiv \inf_{z'\in S}|z - z'|, \quad 
S = \{z \mid \phi'(z) = 0\},
\end{equation}
and reflects the distance of any individual activation to the nearest saturation region. Not only is this penalty simple, but it can be cheaply computed in $O(|\h|)$ time.

\section{Experiments and Results}
\begin{table*}[t]
\caption{Classification accuracies for the various networks. Each column consists of the test set accuracy (left) and the accuracies on the adversarial examples generated from the test set (right). For each model class, the best performance on the adversarial set is in bold.}
\label{t1}
\vskip 0.15in
\begin{center}
\begin{small}
\begin{sc}
\begin{tabular}{|l|c|c|c|r|}
\hline
\abovespace
Training & {Sigmoid MLP} & {ReLU MLP} & {CNN} \\
\hline
\abovespace
Vanilla  & 97.6\%\hspace{1.5mm} 0\% & 98.1\%\hspace{1.5mm} 0.41\% & 99.35\%\hspace{1.5mm} 5.62\% \\ \hline
\abovespace
Adversarial &  92.27\%\hspace{1.5mm} 81.71\%&  92.29\%\hspace{1.5mm} 91.04\% & 99.32\%\hspace{1.5mm} 83.83\%\\ \hline
\abovespace
Saturated  & 97.01\%\hspace{1.5mm} \textbf{94.43\%} & 95.24\%\hspace{1.5mm} \textbf{94.59\%} & 99.33\%\hspace{1.5mm} \textbf{98.45\%} \\ 
\hline
\end{tabular}
\end{sc}
\end{small}
\end{center}
\vskip -0.1in
\end{table*}

\begin{table}[t]
\caption{CNN architecture details.}
\label{t2}
\vskip 0.15in
\begin{center}
\begin{small}
\begin{sc}
\begin{tabular}{|l|r|}
\hline
\abovespace\belowspace
Layer Type & {Architecture} \\ \hline
ReLU Convolutional & 32 filters ($5\times 5$)\\ \hline
Max Pooling & $2\times 2$\\ \hline
ReLU Convolutional &  64 filters ($5\times 5$)\\ \hline
Max Pooling & $2\times 2$\\ \hline
ReLU Fully Connected & 1024 units\\ \hline
Softmax & 10 units\\ \hline
\end{tabular}
\end{sc}
\end{small}
\end{center}
\vskip -0.1in
\end{table}

Here we adapt the above regularization, originally designed for autoencoders, to protect against adversarial examples in supervised classification networks. We found that applying this regularization to every network layer, including the readout layer prior to the softmax output, worked best against adversarial examples generated by the fast gradient sign method. Thus, our penalty took the following form:
\begin{equation}\label{spenalty}
\lambda\sum_{l = 1}^{D}\sum_{i = 1}^{N_l}\phi_c({\h^l_i}).
\end{equation}

Observe that for a ReLU function, the complementary function in \eqref{comp} is itself, so $\phi_c(z) = \max\{0, z\}$. While the definition in \eqref{comp} can also be intricately extended to differentiable functions (as is done in \cite{saturate}), for a sigmoid function we can simply take $\phi_c(z) = |\sigma'(z)| = |\sigma(z)(1-\sigma(z))|$, since the sigmoid is a monotonic function.

We used TensorFlow for all of our models \cite{tensorflow2015-whitepaper}, and we trained both 3 layer multilayer perceptrons (MLPs) with sigmoid and ReLU nonlinearities, as well as convolutional neural networks (CNNs) on 10-class MNIST. For comparison, we trained the adversarially trained networks as in \eqref{eqadv}, finding that $\alpha = 0.5$ gave the best performance. Each network was optimized for performance separately, and we varied the number of hidden units for the MLPs to be between 200-2000 to choose the architecture that provided the best performance. Our CNN architecture is detailed in Table~\ref{t2}, and we used the stronger penalty $f(z) = z$ only at the last layer of the CNN. We used Adam \cite{adam} as our optimizer.

In order to effectively train with the saturating penalty in \eqref{spenalty}, we found that annealing $\lambda$ during training was essential. Starting with $\lambda_{min} = 0$, this was progressively increased to $\lambda_{max} = 1.74$ in steps of size 0.001 for the sigmoidal MLP, $\lambda_{max} = 3.99\times 10^{-8}$ in steps of size $10^{-10}$ for the ReLU MLP, and $\lambda_{max} = 10^{-5}$ in steps of size $10^{-5}$ for the CNN. We ultimately found that the CNN was easier to find an annealing schedule for than the MLPs, further suggesting the viability of this approach in practice.

We list above our results in Table~\ref{t1}. As can be seen, for each model class, we are able to maintain (with little degradation) the original test set accuracy of the network's vanilla counterpart, while also outperforming the adversarially trained counterpart on the adversarial set generated from the test set. We now turn to analyzing the source of adversarial robustness in our networks.


\section{Internal Representation Analysis}
We now examine the internal representations learned by saturating networks (in particular the MLPs) and compare them to those learned by their vanilla counterparts, to gain insight into distinguishing features that make saturating networks intrinsically robust to adversarial examples. 

\begin{figure}[ht]
\vskip 0.2in
\begin{center}
\centerline{\includegraphics[scale = 0.26]{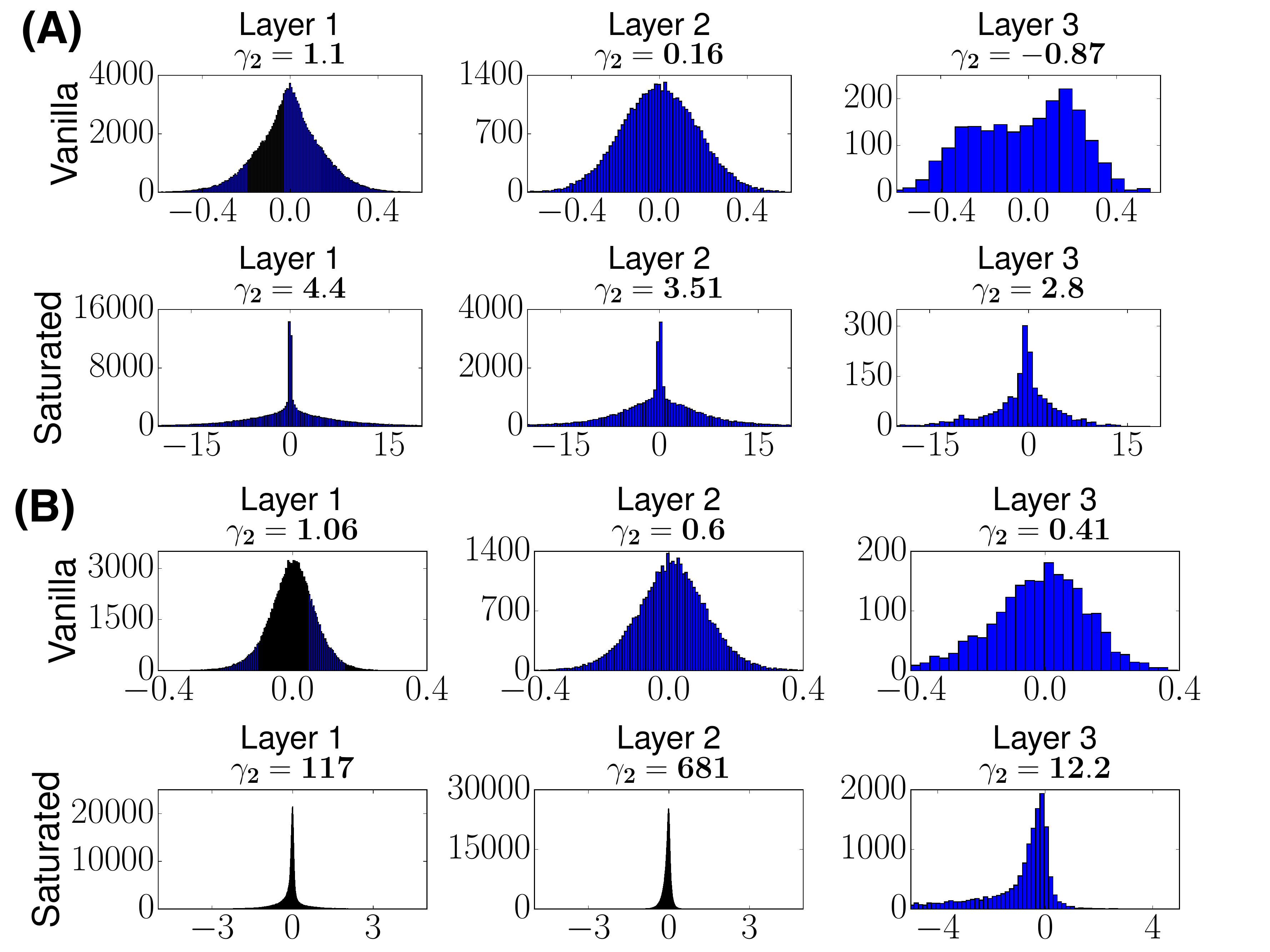}}
\caption{Comparing the vanilla and saturating network's weights for the Sigmoid MLP (A) and ReLU MLP (B) in each layer. The excess kurtosis ($\gamma_2$) of each weight distribution is given as well.}
\label{weights}
\end{center}
\vskip -0.2in
\end{figure} 

\begin{figure}[ht]
\vskip 0.2in
\begin{center}
\centerline{\includegraphics[width=\columnwidth]{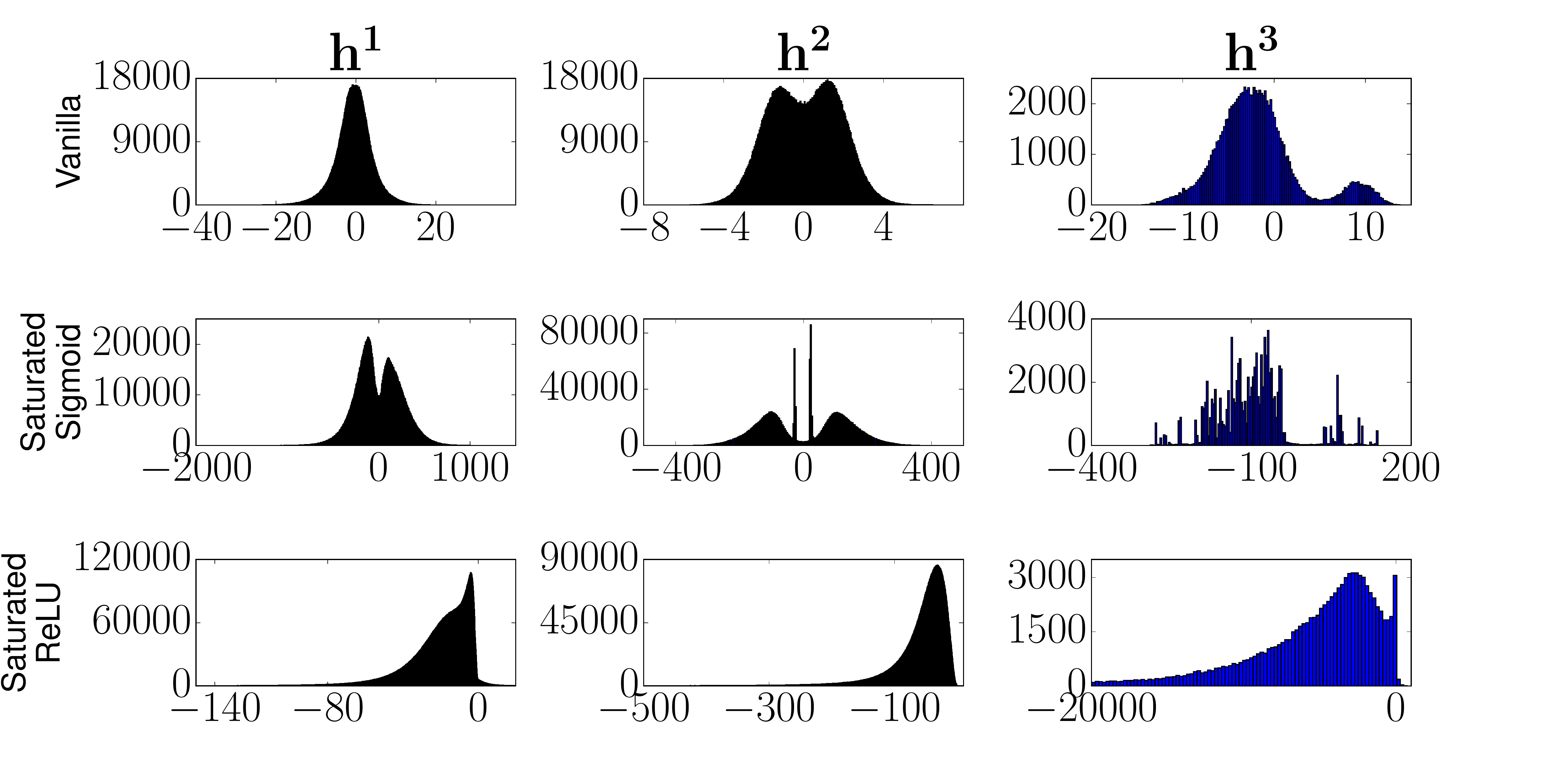}}
\caption{Comparing the vanilla (first row) and saturating network's activities, $\h^{l}$, for the Sigmoid (second row) and ReLU MLP (third row) in each layer $l$.}
\label{activities}
\end{center}
\vskip -0.2in
\end{figure} 

In Figure~\ref{weights}, we compare the weight distributions of the vanilla MLP to the saturating MLP. The saturating MLP weights take on values in a larger range, with a tail that tends to extreme values in the saturating regime of the nonlinearity.  For the sigmoid, this leads to extreme weight values on both ends, while for the saturating ReLU MLP, this leads to extreme negative values. A particularly dramatic change in the weight distribution is a much larger positive excess kurtosis for saturating versus vanilla networks. Indeed, high kurtosis is a property shared by weight distributions in biological networks \cite{lognormal}, raising the question of whether or not it plays a functional role in protection against adversarial examples. In \S 8, we will demonstrate that highly kurtotic weight distributions can act as a linear mechanism to protect against adversarial examples, in addition to the nonlinear mechanism of saturation.

Moreover, in Figure~\ref{activities}, we see that the pre-nonlinearity activations at each layer across all 10,000 test examples also tend to extreme values, as expected, validating that these models are indeed operating in the saturating regime of their respective nonlinearities.

Beyond examining the weights and the activations, we also examine the global structure of internal representations by constructing, for each network and layer, the representational dissimilarity matrix (RDM) of its activities \cite{rdm}. For each of the 10 classes, we chose 100 test set examples at random, and computed the pairwise squared distance matrix,
\begin{equation}
d(\phi\left({\h}^{l,a}\right), \phi\left({\h}^{l,b}\right)) = \frac{1}{N_l}\sum_{i=1}^{N_l} 
\left(\phi\left({\h}^{l,a}_i\right) - \phi\left({\h}^{l,b}_i\right)\right)^2,
\end{equation}
between all pairs $a$ and $b$ of the 1000 test examples. Here $\h^{l,a}$ and $\h^{l,b}$ are the hidden unit activations at layer $l$ on inputs ${\x}^{0,a}$ and ${\x}^{0,b}$, respectively. 

\begin{figure}[ht]
\vskip 0.2in
\begin{center}
\centerline{\includegraphics[scale=0.16]{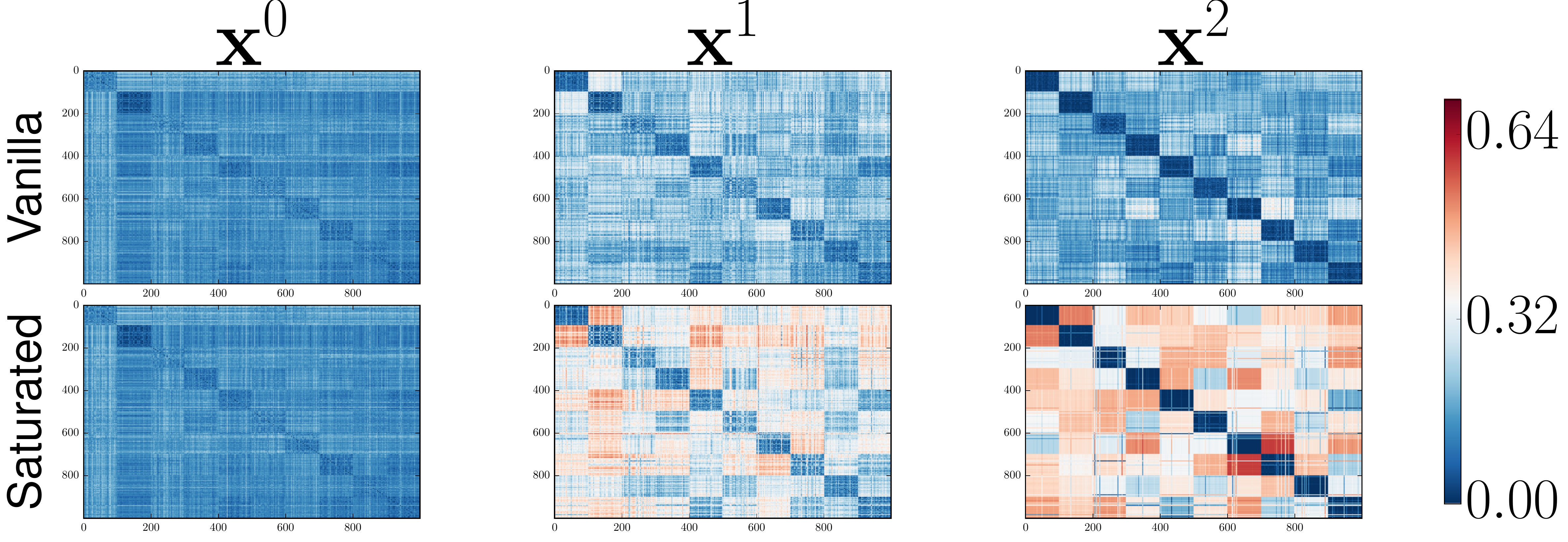}}
\caption{Relational dissimilarity matrix for the Sigmoid MLP. For the corresponding figure for the ReLU MLP, refer to Figure~\ref{relu-rdm} in the Supplementary Material (SM).}
\label{rdm_vis}
\end{center}
\vskip -0.2in
\end{figure} 

As shown in Figure~\ref{rdm_vis}, a distinguishing feature emerges between the RDMs of the vanilla network and the saturated network. At every layer, while within class dissimilarity, in the diagonal blocks, is close to zero for both networks, between class dissimilarities in the off-diagonal blocks are much larger in the saturated network than in the vanilla network. Moreover, this dissimilarity is progressively enhanced in saturating networks as one traverses deeper into the network towards the final output. Thus while both networks form internal representations in which images from each class are mapped to tight clusters in internal representation space, these internal clusters are much further apart from each other for saturating networks. This increased cluster separation likely contributes to adversarial robustness because it necessitates larger norm input perturbations to move representations in deeper layers across decision boundaries, not only in the output layer, but also in intermediate layers.  

\section{The geometry of saturating networks}

While the RDM analysis above showed increased cluster separation in internal representations, we would like to understand better the geometry of the network input-output map and how it contributes to adversarial robustness. To this end, we seek to understand how motions in input space are transformed to motions in the output space of probability distributions over class labels. 

To do so, we rely on the framework of information geometry and Fisher information \cite{amari}. In particular, the network output, as a probability distribution over class labels, is endowed with a natural Riemannian metric, given by the Fisher information.  We can think of the 10 dimensional vector of inputs $\h^D$ in the final layer, as coordinates on this space of distributions (modulo the irrelevant global scaling $\h^D \rightarrow \lambda \h^D$).  In terms of these coordinates, the actual probabilities are determined through the softmax function: $p_i(\h^D) = \frac{1}{Z} e^{h^D_i}$ where $Z = \sum_i e^{h^D_i}$. The Fisher information metric on the space $h^D_i$ is then given by 
\begin{equation}
\begin{split}
{G}^F_{ij} &= \sum_{k}p_k \left(\partial_{z_i}\log p_k\right)\left(\partial_{z_j}\log p_k\right)\\
&= p_i\delta_{ij} - p_ip_j.
\end{split}
\end{equation}
In turn, this metric on $\h^D$ induces a metric on input space $\x^0$ via the pullback operation on metrics. The resultant metric $\mathbf{G}^{in}$ on input space is given by
\begin{equation}\label{pmetric}
\mathbf{G}^{in} = \mathbf{J}^T \mathbf{G}^F \mathbf{J},
\end{equation}
where $\mathbf{J}=\frac{\partial \h^D}{\partial \x^0}$ is the Jacobian from input space to layer $D$. Geometrically, if one moves a small amount from $\x^0$ to $\x^0 + d\x$, the resultant distance $dl$ one moves in output probability space, as measured by the Fisher information metric, is given by 
\begin{equation}
\label{length}
dl = \sqrt{\sum_{ij} G^{in}_{ij} dx_i dx_j}.
\end{equation}
Thus the metric assigns lengths to curves in input space according to how far they induce motions in output space. Also, the Jacobian $\mathbf{J}$ is of independent geometric interest. As a local linearization of the input-output map, the number of non-trivially large singular values of $\mathbf{J}$ determine how many directions in input space the network's input-output map is locally sensitive to. 
\begin{figure}[ht]
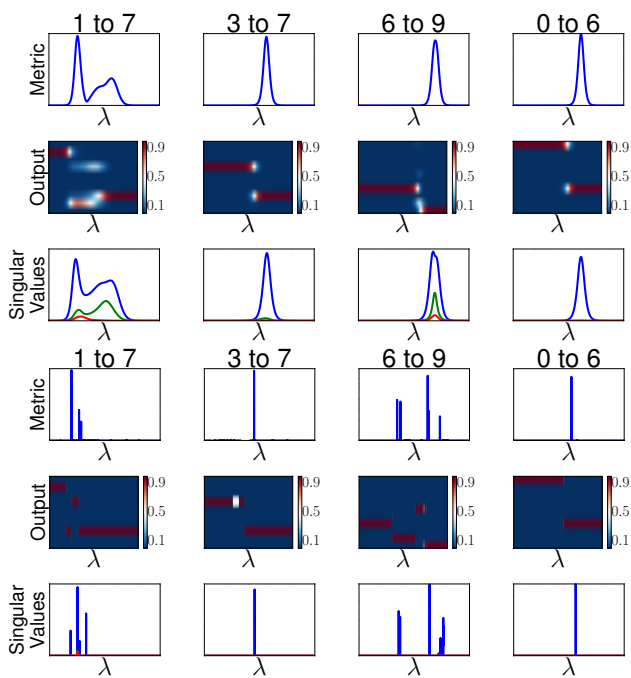

\vskip 0.2in
\begin{center}
\centerline{\includegraphics[width=\columnwidth]{sigmoid_metric_before.pdf}}
\centerline{\includegraphics[width=\columnwidth]{sigmoid_metric_after.pdf}}
\caption{Comparison between the vanilla (top 3 rows) and saturating Sigmoid MLP's (bottom 3 rows) length element, softmax output probabilities, and Jacobian singular values as a function of the interpolation parameter between source and target. The columns denote a particular source and target class pair: $(S: 1, T: 7), (S: 3, T: 7), (S: 6, T: 9), (S: 0, T: 6)$. A similar trend occurs for the ReLU MLP (data not shown).}
\label{metric}
\end{center}
\vskip -0.2in
\end{figure} 

To explore the geometric structure of both vanilla and saturating deep network maps, we move continuously in input space between the most confident images in a given source class, $\x^0_S$, and a target class, $\x^0_T$  along a simple linear interpolation path in pixel space:
\begin{equation}
\x^0(\lambda) = (1-\lambda)\x^0_S + \lambda\x^0_T, \quad \lambda \in [0,1].
\end{equation}
As we move along this path, in Figure~\ref{metric}, we plot the length element in \eqref{length}, the induced trajectory in output probability space, and the spectrum of singular values of the Jacobian $\mathbf{J}$. As expected, the length element increases precisely when the output trajectory in probability space makes large transitions. At these points, one or more singular values of $\mathbf{J}$ also inflate.

Several distinguishing features arise in the geometry of vanilla versus saturated networks in  Figure~\ref{metric}. The first is that the length element is more smooth and continuous for the vanilla network, but locally flat with sharp peaks when class probabilities transition for the saturating network. Thus for saturating networks, one can move long distances in input space without moving much at all in output space. This property likely confers robustness to gradient-based adversaries, which would have difficulty traversing input space under such constant, or flat input-output maps.

A second distinguishing feature is that, in vanilla networks, at probabilistic transition points, multiple singular values inflate, while in saturating networks, only one singular value does so.  This implies that vanilla networks are sensitive to multiple dimensions, while saturating networks perform extremely robust and rapid transitions between distinct probabilistic outputs in a way that sensitivity to input perturbations in {\it all} directions orthogonal to the transition are strongly suppressed. This property again likely confers robustness to adversaries, as it strongly constrains the number of directions of expansion that an adversary can exploit in order to alter output probabilities. 

Finally, it is interesting to compare the geometry of these trained networks to the Riemannian geometry of random neural networks which often arise in initial conditions before training. An extensive analysis of this geometry, performed by \citet{deepchaos}, revealed the existence of two phases in deep networks: a chaotic (ordered) phase when the random weights have high (low) variance. In the chaotic (ordered) phase the network locally expands (contracts) input space everywhere. In contrast, trained networks flexibly deploy both ordered and chaotic phases differentially across input space; they contract space at the center of decision volumes and expand space in the vicinity of decision boundaries. Saturating networks, however, do this in a much more extreme manner than vanilla networks. 

\section{More powerful iterative adversaries}

One can construct more powerful adversaries than the fast sign gradient method by iteratively finding sensitive directions in the input-output map and moving along them. How robust are saturating networks to these types of adversaries? From the information-geometric standpoint described above, given the local flatness of the input-output map, as quantified by our Riemannian geometric analysis, an iterative gradient-based adversary may still encounter difficulty with the saturated network, especially since the number of directions of expansion are additionally constrained by the compressive nature of the map. 

We first created adversaries via iterative first order methods. For each chosen source image $\x_S$ and its associated correct source class $\y_S$, we chose a target class, $\y_T \neq \y_S$. We then attempted to find adversarial perturbations that cause the network to misclassify $\x_s$ as belonging to class $\y_T$. To do so, starting from $\x^{(0)}_{adv}=\x_S$, we iteratively minimized the cross entropy loss $\ell$ via gradient descent:
\begin{equation*}
{\x^{(t+1)}_{adv}} = {\x^{(t)}_{adv}} - \alpha_t\nabla_{\x^{(t)}_{adv}}\ell(\x^{(t)}_{adv}, {\y}_T).
\end{equation*}
This procedure adjusts the adversarial example $\x^{(t)}_{adv}$ so as to make the incorrect label $\y_T$ more likely.  We used Adam \cite{adam} so that the learning rate $\alpha_t$ would be adaptive at each iteration $t$. For a given source class, we started with the source image the network was either least or most confident on.

Although we were able to get the vanilla network to misclassify in either case (usually within less than 10 iterations), there were several cases (such as when the source class was a 3 and the target class was a 7) where we were unable to get the saturated network to misclassify, even in the most extreme case where we ran Adam for {\it 30 million} iterations. Although the image was changing at each iteration and the mean pixel distance from the starting image was steadily increasing and converged, the resultant image did not cause the saturated network to misclassify.

\begin{figure}[ht]
\vskip 0.1in
\begin{center}
\centerline{\includegraphics[width=\columnwidth]{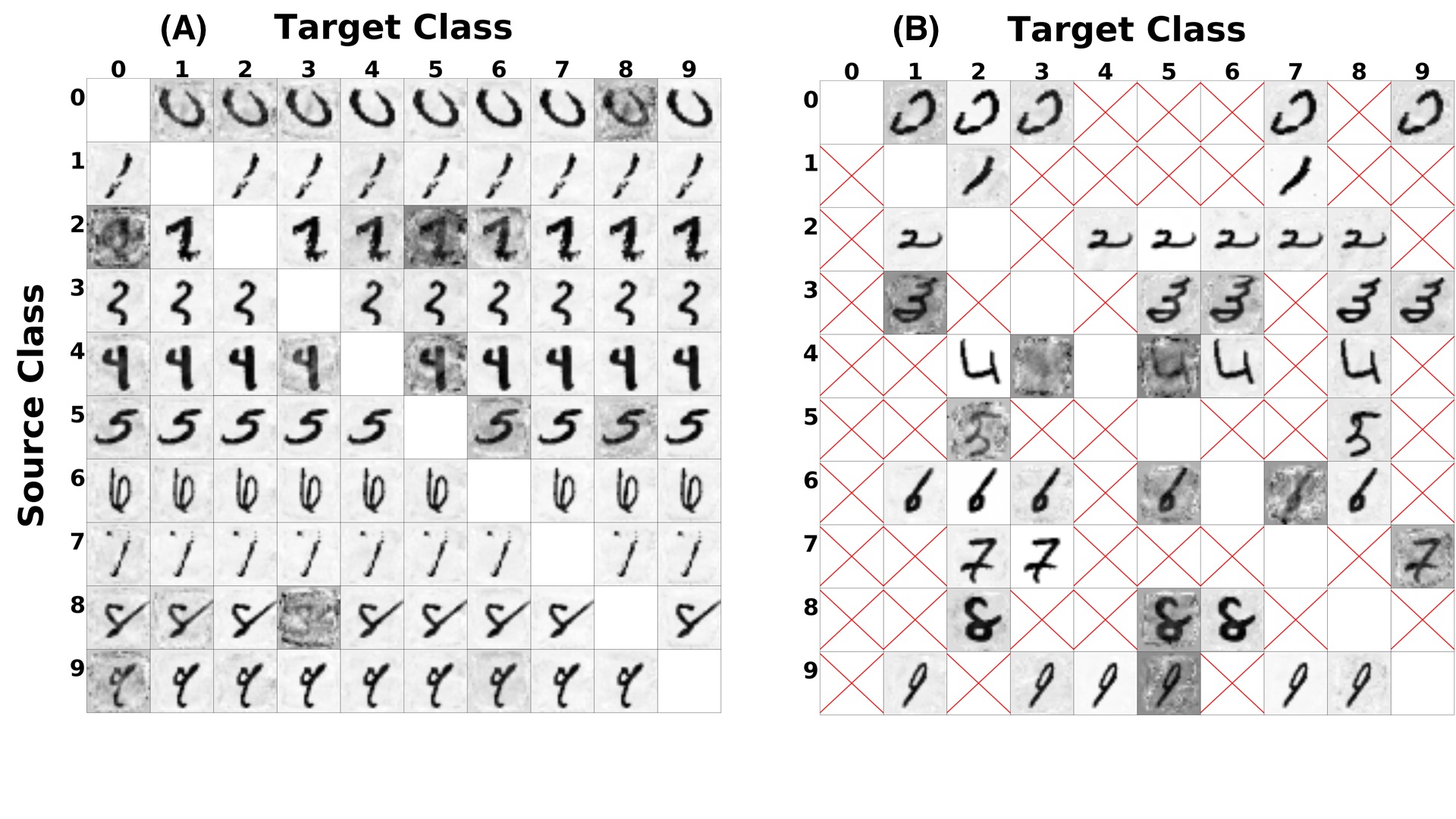}}
\caption{Comparison between the vanilla (A) and saturating Sigmoid MLP's (B) adversarial images after 1000 iterations of L-BFGS. The red `X' denotes that an adversarial image that causes misclassification was not found within 1000 iterations of L-BFGS. The source image was for the least confident image for that source class. A similar trend occurs for the ReLU MLP (data not shown).}
\label{bfgs}
\end{center}
\vskip -0.1in
\end{figure} 

\begin{figure}[ht]
\vskip 0.1in
\begin{center}
\centerline{\includegraphics[scale=0.19]{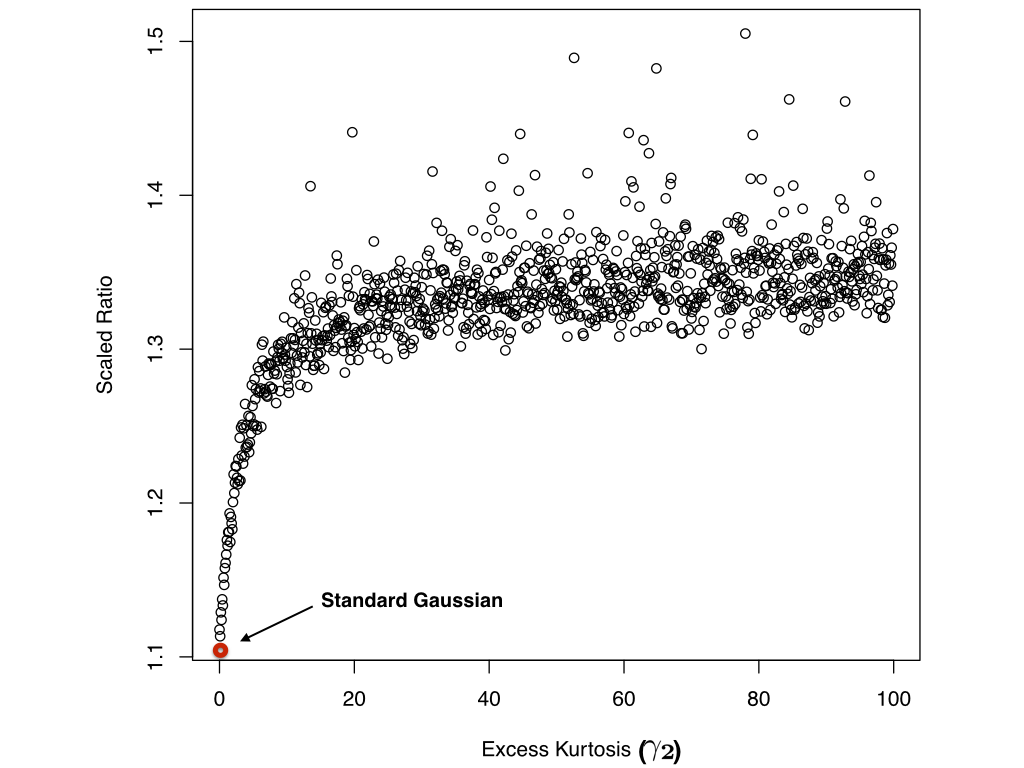}}
\caption{The ratio $\|\w_1\|_2^2/\|\w_2 - \w_1\|_1$, scaled by the mean per component average value of $|\w_1|$, as a function of excess kurtosis ($\gamma_2$) of a Pearson Type VII distribution, which the weights were sampled from. The red point indicates the case of standard Gaussian weights with excess kurtosis 0. Because we are scaling by the intensity, or per-component average of $\w_1$, a value of 1.4 on the y-axis here indicates that the typical per-component value of the perturbation is 1.4 times {\it larger} than the typical per-component strength of the test example.  Thus these perturbations are unlikely to be imperceptible.}
\label{kurtosis}
\end{center}
\vskip -0.1in
\end{figure}

As a result, we moved onto second order adversaries, as \citet{szegedy} had similarly considered. Thus, we considered quasi-Newton methods such as L-BFGS where we would minimize the cross entropy loss $\ell$ as follows:
\begin{equation*}
{\x^{(t+1)}_{adv}} = {\x^{(t)}_{adv}} - \alpha_t{B^{-1}_t}\nabla_{\x^{(t)}_{adv}}\ell(\x^{(t)}_{adv}, {\y}_T),
\end{equation*}
where $B_t$ is the approximate Hessian at iteration $t$ and the learning rate $\alpha_t$ is obtained by performing a line search in the direction $\mathbf{p}_t$ where $B_t\mathbf{p}_t = -\nabla_{\x^{(t)}_{adv}}\ell({\y}_S, {\y}_T)$.

In Figure~\ref{bfgs}, we ran L-BFGS for 1000 iterations on both the vanilla network and the saturated network, starting with a source image that each network correctly classified but had the lowest softmax probability in that class (lowest confidence). In Figure~\ref{kurtosis} in the Supplementary Material (SM), we also include the same analysis, but starting with the most confident source image in each class.

Regardless of whether we start with a source image with the least or highest confidence in that class, we can always find an adversarial image to fool the vanilla network to misclassify as the intended target class (and usually within 1-2 iterations). However, for the saturated network, even starting with the least confident source image, we were, in the majority of cases, unable to fool the network. Moreover, as depicted in Figure~\ref{kurtosis} in the Supplementary Material (SM), it was even more difficult to fool the saturated network with the most confident source image, resulting in only 5 such cases, even after 1000 iterations.

\section{Role of weight kurtosis: a linear mechanism for robustness to adversaries}

As we observed in \S 5, saturating networks had high kurtosis weight distributions in every layer when compared to their vanilla counterparts. Indeed, such kurtotic weight distributions are prevalent in biological neural networks \cite{lognormal}. Here we demonstrate that high kurtosis weight distributions can act as a linear mechanism to guard against adversarial examples.

Indeed, sensitivity to adversarial examples is not unique to neural networks, but arises in many machine learning methods, including linear classification.  Consider for example a classification problem with two cluster prototypes with weight vectors $\w_1\in\mathbb{R}^n$ and $\w_2\in\mathbb{R}^n$. For simplicity, we assume $\w_1$ and $\w_2$ lie in orthogonal directions, so $\w_1 \cdot \w_2=0$. An input $\x$ is classified as class 1 if $\w_1 \cdot \x > 0$, otherwise $\x$ is classified as class $2$. Now consider a test example that is the cluster prototype for class 1, i.e. $\x = \w_1$. Let us further consider an adversarial perturbation $\w_1 + \Delta\x$. This perturbed input will be misclassified if and only if $(\w_2 - \w_1)\cdot (\w_1 + \Delta\x) > 0$.  Following the fast sign gradient method, we can choose $\Delta\x$ to be the maximum perturbation under the constraint $\|\Delta\x\|_{\infty} < \epsilon$ in \eqref{eq2}. This optimal perturbation is 
$\Delta\x = \epsilon\text{ sgn}(\w_2 - \w_1)$. 
In order to have this bounded $l_\infty$ norm perturbation cause a misclassification, we must then have the condition
\begin{equation}\label{normratio}
\epsilon > \epsilon_{\text{min}}  \equiv \frac{\|\w_1\|_2^2}{\|\w_2 - \w_1\|_1}.
\end{equation}
Here, recall we are assuming $\w_1 \cdot \w_2 = 0$ for simplicity. Thus if the $l_1$ norm in the denominator is small, then the network is adversarially robust in the sense that a large perturbation is required to cause a misclassification, whereas if the $l_1$ norm is large, then it is not. 

Now in high dimensional spaces, $l_1$ norms can be quite large relative to $l_2$ norms. In particular for any unit $l_2$ norm vector $\mathbf{v}$, we have $1 \le \|\mathbf{v}\|_1 \le \sqrt{n}$,
where the upper bound is realized by a dense uniform vector with each entry $\frac{1}{\sqrt{n}}$ and the lower bound is realized by a coordinate vector with one nonzero entry equal to $1$. Both these vectors are on the $l_2$ ball of radius $1$, but this $l_2$ ball intersects the circumscribing $l_1$ ball of radius $\sqrt{n}$ at the former vector, and the  inscribing $l_1$ ball of radius $1$ at the latter vector.  This intersection of $l_1$ and $l_2$ balls of very different radii in a high dimensional space likely contributes to the prevalence of adversarial examples in high dimensional linear classification problems by allowing the denominator in \eqref{normratio} to be large and the numerator to be small.  

However, we can avoid the bad regime of dense uniform vectors with large $l_1$ norm if the weights are sampled from a kurtotic distribution.
In this case, we may then expect that the numerator $\|\w_1\|_2^2$ in \eqref{normratio} would be large as we are likely to sample from extreme values, but that the denominator $\|\w_2 - \w_1\|_1$ would be small due to the peak of the distribution near 0. To test this idea, we sampled unit norm random vectors of dimension 20000, so that $\w_1\cdot\w_2 \approx 0$. We sampled their values iid from a Pearson Type VII distribution, with density function given by
\begin{equation*}
f(x; \gamma_2) = c(\gamma_2)\left(1 + \left(\frac{x}{\sqrt{2+6/\gamma_2}}\right)^2\right)^{-5/2-3/\gamma_2},
\end{equation*}
where $c(\gamma_2) = \frac{1}{\left(\sqrt{2+6/\gamma_2}\right) B\left(2 + 3/\gamma_2, \frac{1}{2}\right)}$, $B$ is the Euler Beta function, and $\gamma_2$ denotes the excess kurtosis of the distribution. In Figure~\ref{kurtosis}, we computed the ratio in \eqref{normratio} and scaled it by the input intensity, which is given by the average absolute value of a nonzero component of $\w_1$. The resultant scaled ratio was then computed for each value of $\gamma_2$. Note that a standard Gaussian has an excess kurtosis of 0, which serves as a baseline. Hence, increasing the excess kurtosis via, for example, a Pearson Type VII density, increases the scaled ratio by almost $40\%$ from the Gaussian baseline. In fact, if we sample, via inverse transform sampling, from the weight distribution of the saturated network at a given layer, then the scaled ratio can increase by as much as $300\%$ from when we sample from the distribution of the weights in that same layer for the vanilla network. Thus even in the case of linear classification, kurtotic weight distributions, including the weight distributions learned by our saturating networks, can improve robustness to adversarial examples. 

\section{Discussion}
In summary, we have shown that a simple, biologically inspired strategy for finding highly nonlinear networks operating in a saturated regime provides interesting mechanisms for guarding DNNs against adversarial examples without ever computing them. Not only do we gain improved performance over adversarially trained networks on adversarial examples generated by the fast gradient sign method, but our saturating networks are also relatively robust against iterative, targeted methods including second-order adversaries. We additionally move beyond empirical results to analyze the sources of intrinsic robustness to adversarial perturbations. Our information geometric analyses reveal several important features, including highly flat and low dimensional internal representations that nevertheless widely separate images from different classes.  Moreover, we have demonstrated that the highly kurtotic weight distributions found both in our networks and in our brains, can act as a linear mechanism against adversarial examples. Overall, we hope our results can aid in combining theory and experiment to form the basis of a general theory of biologically plausible mechanisms for adversarial robustness.

\section*{Acknowledgements}
We thank Ben Poole and Niru Maheswaranathan for helpful comments on the manuscript, and the ONR, and Burroughs Welcome, Simons, McKnight, and James. S. McDonnell foundations for support.

\bibliography{paper_bib}
\bibliographystyle{icml2017}

\begin{figure*}[ht]
\vskip 0.2in
\begin{center}
\centerline{\includegraphics[scale = 0.35]{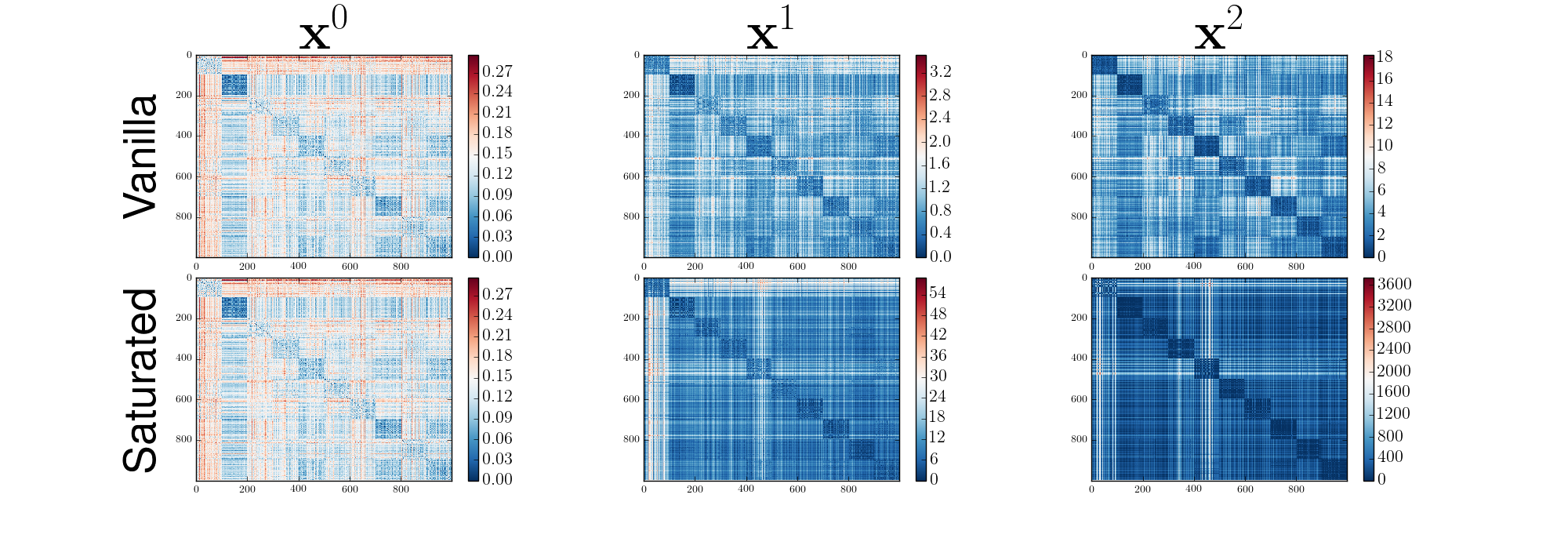}}
\caption{Relational dissimilarity matrix for the ReLU MLP.}
\label{relu-rdm}
\end{center}
\vskip -0.2in
\end{figure*} 

\begin{figure*}[ht]
\vskip 0.2in
\begin{center}
\centerline{\includegraphics[scale = 0.25]{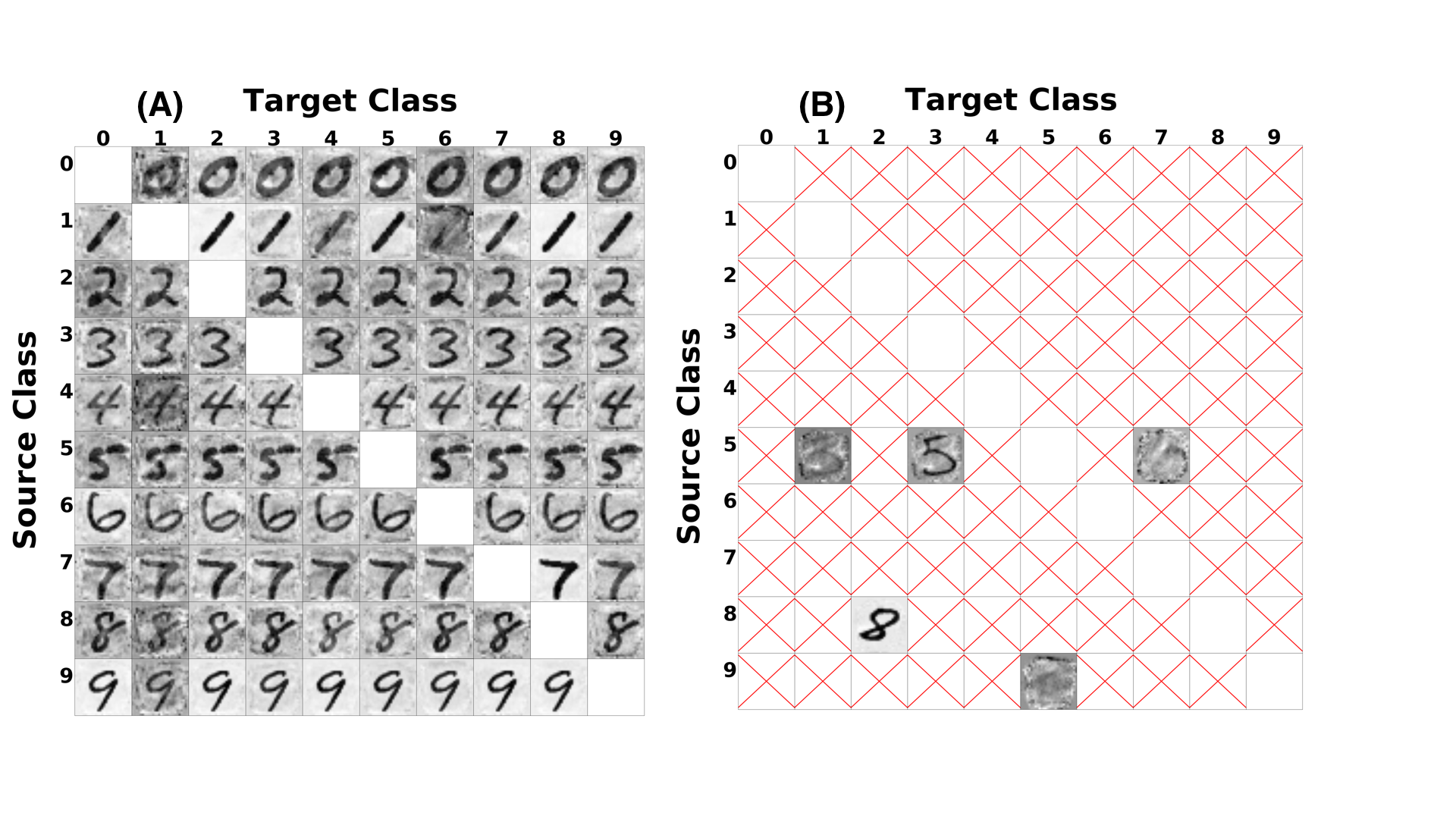}}
\caption{Comparison between the vanilla (A) and saturating Sigmoid MLP's (B) adversarial images after 1000 iterations of L-BFGS. The red `X' denotes that an adversarial image was not found in order to cause the network to misclassify for that target class. The source image was for the most confident image for that source class.}
\label{kurtosis}
\end{center}
\vskip -0.2in
\end{figure*}
\end{document}